\documentclass{article}
\usepackage{iclr2027_conference,times}

\usepackage{amsmath,amsfonts,bm}

\def\eqref#1{equation~\ref{#1}}

\def\1{\bm{1}}

\DeclareMathAlphabet{\mathsfit}{\encodingdefault}{\sfdefault}{m}{sl}
\SetMathAlphabet{\mathsfit}{bold}{\encodingdefault}{\sfdefault}{bx}{n}

\usepackage{hyperref}
\usepackage{url}
\usepackage{graphicx}
\usepackage{float}
\usepackage{needspace}
\usepackage{algorithm}
\usepackage{algpseudocode}
\usepackage{wrapfig}
\usepackage{capt-of}
\usepackage{booktabs}
\usepackage{multirow}
\usepackage{xcolor}

\newcommand{\best}[1]{\textcolor{red}{\textbf{#1}}}
\newcommand{\second}[1]{\textcolor{blue}{\underline{#1}}}

\title{ReCaVSR: One-Step Streaming Diffusion Video Super-Resolution with Recycled Latents and Learned Cache Routing}
\author{\parbox[t]{\dimexpr\textwidth-2\tabcolsep\relax}{\centering
\textbf{Xijun Wang$^{1}$ \quad Xin Li$^{1}$ \quad
Suhang Yao$^{1}$ \quad Zirui Lang$^{1}$ \quad Bingchen Li$^{1}$} \\
\textbf{Zhibo Chen$^{1}$} \\
\vspace{5pt}
{\normalfont $^{1}$University of Science and Technology of China} \\
}
}

\iclrfinalcopy
\begin{document}
\maketitle
\lhead{Preprint}

\begin{abstract}
Real-time diffusion-based video super-resolution (VSR) is in high demand for online streaming,
yet stringent latency requirements often compromise generative fidelity.
We propose \emph{ReCaVSR}, a Wan2.2-based, one-step framework for streaming
VSR that builds on two observations: recycled SR latents retain local temporal
context, reducing the need for full historical Key-Value (KV) caches; and
individual transformer layers benefit from distinct temporal scopes.
ReCaVSR combines three complementary designs:
\textit{(i)} layer-wise cache routing with recycled SR latents:
each DiT layer learns its KV-cache temporal scope under a cache budget
and exports a static inference schedule,
while recycled SR latents propagate local context by conditioning each new block
on the model's own preceding predictions.
\textit{(ii)} Multi-Scope Query (MSQ) Discriminator:
a compositional discriminator combining global, spatial-window, and temporal-tube feedback
for holistic realism, local texture generation, and temporal stability.
\textit{(iii)} LR-conditioned adaptation of FlashDecoder:
a VAE decoder that incorporates LR observations for efficient latent decoding.
ReCaVSR enables streaming VSR without iterative sampling
or full historical KV-cache materialization.
Experiments on synthetic and real-world VSR benchmarks
show better perceptual quality, temporal consistency, and streaming efficiency
than representative VSR baselines.
At $1080{\times}1920$ output resolution on a single
NVIDIA A100-80GB, ReCaVSR achieves 21.20 FPS with 15.16 GB
peak allocated GPU memory, running 2.72$\times$ faster while using
38.0\% less peak allocated memory than FlashVSR Tiny.
The code is available at \url{https://github.com/kopperx/ReCaVSR}.

\end{abstract}

\section{Introduction} \label{sec:intro}

Video super-resolution (VSR) recovers high-resolution videos from
low-resolution inputs and is important for latency-sensitive applications
such as video streaming and telepresence.
We study \emph{one-step streaming VSR}, where each high-resolution latent
block is generated in a single forward pass using only current and past
input blocks.

Recurrent and transformer-based VSR methods exploit temporal information,
but their perceptual details can remain limited under severe degradation~\citep{shiu2025stream}.
Diffusion-based methods improve texture realism through strong generative priors,
but iterative sampling and offline temporal modeling increase latency~\citep{chen2025dove}.
Recent one-step and streaming variants alleviate these costs, but leave
the efficient use of temporal context unresolved~\citep{zhuang2025flashvsr}.
Autoregressive video generation typically reuses historical Key-Value (KV)
states for causal generation~\citep{huang2025self}; in VSR, however,
current LR observations and previously generated SR latents already provide
strong local temporal cues.
Materializing full historical KV caches in every diffusion transformer layer
is therefore unnecessarily costly in computation and memory.

\begin{wrapfigure}{r}{0.54\textwidth}
    \vspace{-8pt}
    \centering
    \includegraphics[width=\linewidth]{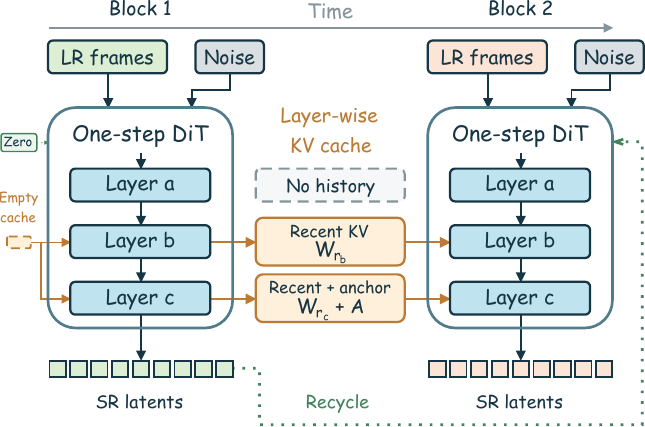}
    \vspace{-6pt}
    \caption{\textbf{Streaming inference with ReCaVSR.}
    Recycled SR latents convey local temporal context from the preceding block,
    while layer-wise cache routing retains only the historical KV states required
    by each DiT layer.}
    \label{fig:streaming_intro}
    \vspace{-8pt}
\end{wrapfigure}

ReCaVSR is built on two key insights about streaming diffusion-based VSR\@.
First, recycling SR latents means feeding the SR latents
from the preceding generated block back to condition the current prediction.
This recurrence preserves sufficient local temporal context,
reducing the need to access the full historical KV cache
in every diffusion transformer layer.
Second, different transformer layers require different temporal scopes,
ranging from no historical cache to partial caches of recent frames,
keyframe anchors, or their combination.
These insights suggest that recycled SR latents should provide local temporal propagation,
while historical K/V should be materialized only where each layer needs longer-range memory,
as illustrated in Figure~\ref{fig:streaming_intro}.

In this work, we propose \emph{ReCaVSR},
a Wan2.2-based one-step streaming VSR framework.
The first complementary design is layer-wise cache routing with recycled SR latents.
The generator receives causal LR evidence from a causal LR projector
and uses recycled SR latents as a recurrent high-resolution condition.
During Stage~1, we use teacher forcing: clean HQ latents from the preceding
block provide the recurrent condition while adapting the generator for VSR.
During the later phase of Stage~1, the cache router learns a KV-cache temporal
scope for each DiT layer under a fixed cache budget,
choosing among no history, recent-frame windows, keyframe anchors,
and their combination.
The learned routing distribution is sharpened and exported as a static schedule
for Stage~2 and streaming inference,
so streaming inference keeps only the historical K/V selected by each layer.

The second design is the Multi-Scope Query (MSQ) Discriminator
for VSR-aware adversarial one-step training.
Following AAPT~\citep{lin2025autoregressive}, APT~\citep{lin2025diffusion},
and recent one-step restoration methods~\citep{wang2025seedvr2,chen2025dove,lv2026duo,chen2026improved},
the generator is trained for one-step perceptual generation.
However, an APT-style global-query discriminator compresses a video clip
into a single holistic real/fake judgment,
which can miss failure modes that are local in space or time.
The MSQ Discriminator is therefore a compositional discriminator
with global, spatial-window, and temporal-tube query groups.
These scopes provide adversarial feedback for holistic realism,
local texture generation, and temporal stability,
targeting the spatial detail and flicker artifacts that are common in one-step VSR\@.
During Stage~2, we use sequential self-rollout with recycled SR latents under
the exported cache schedule.
Each block is generated in one step conditioned on the recycled SR latents from
the model's preceding predictions~\citep{huang2025self,guo2025end}.

The third design adapts FlashDecoder~\citep{kang2026flashdecoder}
for LR-conditioned latent decoding.
Upsampled LR frames are projected onto the latent spatial grid
and injected at two complementary stages:
grouped LR features condition the Transformer backbone,
whereas frame-aligned features guide temporal refinement after temporal upsampling.
This design provides direct access to LR observations during decoding
while preserving FlashDecoder's rolling KV cache
and introducing no additional attention tokens.

Our experiments evaluate both VSR quality and streaming behavior.
Across synthetic and real-world VSR benchmarks, ReCaVSR achieves strong
perceptual quality and temporal consistency relative to representative VSR
baselines while maintaining competitive distortion quality.
At $1080{\times}1920$ output resolution on a single NVIDIA A100-80GB,
ReCaVSR runs at 21.20 FPS with 14.12 GiB peak allocated GPU memory.
Compared with FlashVSR Tiny,
it is 2.72$\times$ faster while using 38.0\% less peak allocated memory.
The main contributions are summarized as follows:

\begin{itemize}
    \item We introduce ReCaVSR,
    a Wan2.2-based one-step streaming VSR framework
    that combines sequential self-rollout training with an LR-conditioned
    adaptation of FlashDecoder for efficient latent decoding.

    \item We propose layer-wise cache routing with recycled SR latents,
    which couples local temporal propagation through recycled SR latents
    with learned, layer-specific KV-cache scopes under a cache budget,
    and exports them as a static schedule for streaming inference.

    \item We introduce the Multi-Scope Query (MSQ) Discriminator,
    which combines global, spatial-window, and temporal-tube queries
    to provide adversarial feedback at complementary spatial and temporal scopes.
\end{itemize}

\section{Related Work} \label{sec:relate}

\subsection{Real-world video super-resolution}

Video super-resolution (VSR) has traditionally exploited temporal redundancy
through alignment, propagation, and aggregation~\citep{wang2019edvr,chan2021basicvsr,liang2022recurrent}.
Real-world methods further investigate how complex degradations affect temporal
propagation and attention-based reconstruction~\citep{chan2022investigating,zhang2024realviformer}.
Generative approaches introduce adversarial or pretrained diffusion priors
to synthesize realistic details while maintaining temporal coherence~\citep{xu2025videogigagan,zhou2024upscale,li2025diffvsr}.
More recent frameworks, including SeedVR and Vivid-VR, adapt video diffusion
transformers to degraded observations through window-based modeling and
improved conditional control~\citep{wang2025seedvr,bai2026vividvr}.
To reduce iterative sampling costs, one-step VSR methods explore latent-pixel
adaptation, diffusion distillation, and adversarial post-training~\citep{chen2025dove,wang2025seedvr2,lv2026duo}.
Building on this line of work, ReCaVSR introduces the MSQ Discriminator,
which combines global, spatial-window, and temporal-tube queries to provide
adversarial feedback for holistic realism, local texture generation, and
temporal stability in one-step VSR.

\subsection{Streaming video super-resolution and generation}

Streaming video super-resolution and generation require temporal continuity
under incremental processing.
Autoregressive video diffusion models use causal attention and KV caching
to reuse historical context~\citep{yin2025slow,lin2025autoregressive}.
Student-forcing and self-forcing training further expose models to their own
generated histories, aligning the training process with autoregressive
inference~\citep{lin2025autoregressive,huang2025self}.
In VSR, FlashVSR combines locality-constrained sparse attention with parallel
one-step distillation, while InfVSR maintains temporal context through rolling
KV caches and LR visual guidance~\citep{zhuang2025flashvsr,zhang2026infvsr}.
SwiftVR instead processes each latent chunk without a rolling DiT KV cache,
maintaining cross-chunk continuity through its streaming autoencoder~\citep{yan2026swiftvr}.
Stream-DiffVSR follows a strictly frame-by-frame protocol with four-step diffusion
and autoregressive guidance from aligned previous SR frames~\citep{shiu2025stream}.
ReCaVSR separates local and longer-range temporal information by combining
recycled SR latents for local temporal propagation with layer-specific
historical KV states for longer-range context.

\subsection{Efficient video diffusion attention}

Efficient attention is central to applying video diffusion transformers
to long sequences and high resolutions.
Feature-caching methods reuse intermediate activations across denoising steps
to reduce repeated computation~\citep{zhao2408real,lv2024fastercache},
while sparse and hybrid attention operators reduce the cost of token interactions
within each step~\citep{zhang2025vsa,zhang2025sla}.
In VSR, FlashVSR employs locality-constrained sparse attention, and TRaM-VSR
routes and merges video tokens within selected network-depth
intervals~\citep{zhuang2025flashvsr,gao2026tramvsr}.
For autoregressive generation, selecting historical K/V states provides another
way to reduce attention cost by limiting the context accessed at each generation step.
HeadCast assigns attention heads different cache pathways through
training-free profiling~\citep{shen2026headcast}.
Unlike feature caching across denoising steps or token pruning within a step,
ReCaVSR allocates temporal history across DiT layers.
It learns a budget-constrained KV-cache scope for each layer and exports the
result as a static schedule so that only selected historical states are stored
and attended to during streaming inference, reducing KV-cache memory and
historical-attention computation.

\section{Method} \label{sec:method}

\begin{figure}[t]
\centering
\includegraphics[width=\linewidth]{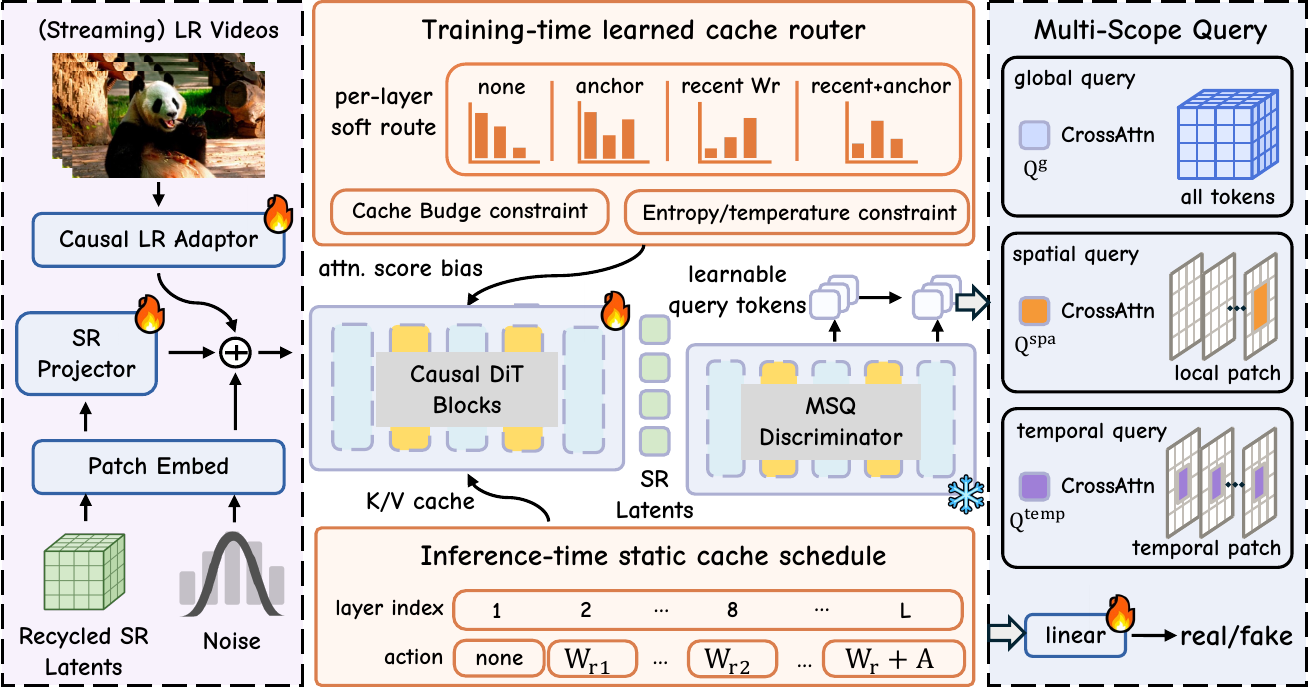}
\setlength{\abovecaptionskip}{0pt}
\caption{\textbf{Overview of ReCaVSR\@.}
The generator conditions on LR observations and recycled SR latents,
with layer-wise routing selecting historical KV states.
In Stage~1, cache allocation is learned under budget and entropy regularization
and exported as a static per-layer schedule.
In Stage~2, the generator performs one-step sequential rollout under this schedule,
supervised by the MSQ discriminator through global, spatial-window, and temporal-tube queries.}
\label{fig:method_overview}
\vspace{-4pt}
\end{figure}

\vspace{-2pt}
\paragraph{Overview.}
ReCaVSR is a Wan2.2-based one-step streaming video super-resolution (VSR) framework,
illustrated in Fig.~\ref{fig:method_overview}.
Given a low-resolution video stream,
we denote the LR inputs aligned with generation block $k$ by $x_k^{LR}$
and sequentially predict the corresponding SR latent block:
\begin{equation}
    \widehat{z}_k^{SR}
    =G_{\theta}\!\left(
    \epsilon_k,x_k^{LR},\widehat{z}_{k-1}^{SR},\mathcal{M}_{<k}
    \right),
    \qquad
    \epsilon_k\sim\mathcal{N}(0,I),
    \label{eq:overall_generator}
\end{equation}
where $\widehat{z}_{k-1}^{SR}$ denotes the preceding generated SR latent block
and $\mathcal{M}_{<k}$ denotes the historical KV cache.
Latent recycling supplies local temporal conditioning,
and learned layer-wise routing selects the additional KV history.
An LR-conditioned FlashDecoder converts the predicted latents into RGB frames.
Training proceeds in two stages:
VSR adaptation with layer-wise cache learning (Stage~1, Sec.~\ref{sec:routed_cache}),
followed by one-step adversarial post-training under the exported cache schedule
(Stage~2, Sec.~\ref{sec:adv_training}).
Section~\ref{sec:lr_flashdecoder} describes the LR-conditioned FlashDecoder
used for latent decoding.

\subsection{Recycled Latents and Layer-Wise Cache Routing}
\label{sec:routed_cache}

We project the preceding SR latent block into the generator's hidden dimension
and add it to the current latent features together with the LR conditioning features.
With this recurrent condition providing the latest SR state,
we learn each transformer layer's access to additional KV history under a shared cache budget.

\vspace{-2pt}
\paragraph{Cache allocation.}
Each layer selects historical KV states from a recent bank and a sparse anchor bank.
The candidate set $\mathcal{A}$ includes no historical access ($\emptyset$),
recent windows ($W_r$), anchors ($A$), and combined actions ($W_r+A$).
$W_r$ contains the latest $r$ historical latent positions,
while $A$ contains periodically sampled positions with a fixed capacity $c_A$.
A router predicts action probabilities
$\pi_l=\operatorname{softmax}(R_\eta(e_l)/\tau)$,
where $e_l$ is the embedding of layer $l$, $R_\eta$ is a learnable network,
and $\tau$ is the temperature.
Each distribution is shared across attention heads and depends only on layer identity.

\vspace{-2pt}
\paragraph{Soft routing.}
To learn discrete cache choices through backpropagation,
we relax action selection into a soft visibility prior over historical keys.
During route learning, attention includes the union of candidate historical states.
This allows the router to receive gradients for all candidate actions
within a single attention operation.
For a query $i$ and historical key $j$, let $\mathbf{1}_a(i,j)$
indicate whether action $a$ permits access.
We aggregate the probabilities of all actions that retain this key
and use the resulting visibility weight to bias its attention score:
\begin{equation}
    P_l(i,j)=\sum_{a\in\mathcal{A}}\pi_l(a)\mathbf{1}_a(i,j),
    \quad
    \widetilde{s}_{ij}^{\,l}
    =s_{ij}^{\,l}+\log\!\left(P_l(i,j)+\delta\right),
    \label{eq:routed_score}
\end{equation}
where $s_{ij}^{\,l}$ is the original attention score
and $\delta$ is a small numerical constant.
Adding the log prior scales the unnormalized attention weight
$\exp(s_{ij}^{\,l})$ by $P_l(i,j)+\delta$.
Historical keys supported by low-probability actions therefore contribute less to attention,
while current-block scores remain unchanged.
This differentiable weighting allows the VSR loss to guide cache allocation.
The subsequent budget and sharpening terms regulate its capacity
and encourage a discrete choice for export.

\vspace{-2pt}
\paragraph{Training and export.}
Stage~1 first adapts the generator to VSR with fixed cache scopes,
using clean HQ latents as the recurrent condition.
In the later phase, we enable soft routing and jointly optimize the generator and router.
The VSR adaptation objective $\mathcal{L}_{\mathrm{FM}}$
is augmented with a cache-budget term and an entropy penalty:
\begin{equation}
    \begin{gathered}
        \mathcal{L}_{\mathrm{Stage1}}
        =\mathcal{L}_{\mathrm{FM}}
        +\lambda_{\mathrm{budget}}\bigl(\bar{c}-c_{\mathrm{target}}\bigr)^2
        +\frac{\lambda_{\mathrm{sharp}}}{L}\sum_{l=1}^{L}H(\pi_l),\\[-2pt]
        \bar{c}
        =\frac{1}{L}\sum_{l=1}^{L}
        \sum_{a\in\mathcal{A}}\pi_l(a)c(a),
    \end{gathered}
    \label{eq:stage1_objective}
\end{equation}
where $L$ is the number of layers, $H$ is categorical entropy,
and $c(a)$ measures reserved cache capacity in latent positions.
We assign capacities $0$, $r$, $c_A$, and $r+c_A$
to $\emptyset$, $W_r$, $A$, and $W_r+A$, respectively.
The budget term encourages the average expected capacity to approach $c_{\mathrm{target}}$,
while entropy minimization and temperature annealing
concentrate each layer's probability on a dominant action.
At the end of Stage~1, we export
$a_l^*=\arg\max_{a\in\mathcal{A}}\pi_l(a)$ for each layer.
This allocation remains fixed during Stage~2.
Each layer retains and accesses only the selected historical KV states,
without evaluating the router or applying the soft routing bias.

\subsection{Multi-Scope Query Discriminator}
\label{sec:adv_training}

In Stage~2, we optimize the adapted generator through sequential self-rollout
under the exported cache schedule,
combining MSQ adversarial feedback with latent and RGB reconstruction supervision.
One-step VSR requires adversarial feedback on holistic realism,
local texture generation, and temporal stability.
We introduce a Multi-Scope Query (MSQ) Discriminator that addresses these aspects
through global, spatial-window, and temporal-tube query groups.

Following APT~\citep{lin2025diffusion} and AAPT~\citep{lin2025autoregressive},
we use a video-DiT backbone to extract features from real and generated latent clips.
Global queries aggregate features across the full clip
to assess overall appearance and structure.
Spatial-window queries inspect local regions within individual latent frames,
while temporal-tube queries examine the same spatial region across consecutive latent frames.
We average the query logits within each scope
and then equally combine the three scope-level scores into the final discriminator logit.
\vspace{-2pt}
\paragraph{Relativistic adversarial objective.}
We adopt a relativistic standard GAN (RSGAN) objective.
Let $c_r$ and $c_f$ denote the discriminator logits
for paired real and generated latent clips.
The adversarial objectives are
\begin{equation}
    \mathcal{L}_{\mathrm{adv}}^{D}
    =\mathbb{E}\left[\operatorname{softplus}(c_f-c_r)\right],
    \quad
    \mathcal{L}_{\mathrm{adv}}^{G}
    =\mathbb{E}\left[\operatorname{softplus}(c_r-c_f)\right],
    \label{eq:rsgan_objective}
\end{equation}
where $\operatorname{softplus}(u)=\log(1+e^u)$.
During Stage~2, we combine these adversarial objectives with latent and RGB
reconstruction losses for the generator and feature R1 regularization for the discriminator:
\begin{equation}
    \begin{gathered}
        \mathcal{L}_{\mathrm{stage2}}
        =
        \lambda_{\mathrm{FM}}\mathcal{L}_{\mathrm{FM}}
        +\lambda_{\mathrm{adv}}\mathcal{L}_{\mathrm{adv}}^{G}
        +\lambda_{\mathrm{RGB}}\mathcal{L}_{\mathrm{RGB}}
        +\lambda_{\mathrm{perc}}\mathcal{L}_{\mathrm{perc}},\\
        \mathcal{L}_{D}
        =
        \mathcal{L}_{\mathrm{adv}}^{D}
        +\gamma_{\mathrm{fR1}}\mathcal{L}_{\mathrm{fR1}}.
    \end{gathered}
    \label{eq:total_loss}
\end{equation}
Here, $\mathcal{L}_{\mathrm{RGB}}$ and $\mathcal{L}_{\mathrm{perc}}$
are RGB MSE and LPIPS losses computed through the frozen VAE decoder.
To avoid out-of-memory errors from second-order backpropagation through the discriminator backbone,
we design a feature-space R1 regularizer.
It penalizes discriminator-logit gradients with respect to real backbone features.
We detach these features from the backbone for the regularization branch,
restricting second-order differentiation to the MSQ heads.

\subsection{LR-Conditioned FlashDecoder}
\label{sec:lr_flashdecoder}

In our one-step VSR pipeline,
the original Wan VAE decoder becomes the main runtime bottleneck.
We adopt the Transformer architecture of FlashDecoder~\citep{kang2026flashdecoder}
and extend it with LR conditioning, yielding a 57.17M-parameter decoder.
Compared with convolutional decoders that process progressively enlarged feature maps,
this architecture performs spatiotemporal modeling on the low-resolution latent grid
and defers spatial expansion to the final output projection.
A fixed-size rolling KV cache reuses historical features
while keeping per-frame decoding cost bounded as the video grows.

We project upsampled LR frames onto the latent grid and add grouped features
to the Transformer backbone and frame-aligned features to the temporal refinement layers
after temporal upsampling, without adding attention tokens.
We train the decoder independently using HQ latents from a frozen Wan encoder
and paired LR inputs, using $L_1$ and LPIPS losses for RGB reconstruction.

\begin{table}[!b]
\begingroup
\centering
\caption{Quantitative comparison on synthetic and real-world VSR benchmarks.
The best and second performances are marked in \best{red} and \second{blue}, respectively.}
\label{tab:main_results}
\renewcommand{\arraystretch}{1.15}  \resizebox{\textwidth}{!}{
\setlength{\heavyrulewidth}{1.2pt}
\begin{tabular}{cl|ccccccc|c}
\toprule
Dataset & Metric & RealViformer & UAV & STAR & DOVE & SeedVR2 & SwiftVR & FlashVSR & Ours \\
\midrule

\multirow{7}{*}{\textbf{REDS30}}
 & PSNR $\uparrow$ & \best{23.32} & 21.15 & 22.14 & \second{23.28} & 22.17 & 21.33 & 21.41 & 21.67 \\
 & SSIM $\uparrow$ & \second{0.5967} & 0.5143 & 0.5432 & \best{0.6103} & 0.5860 & 0.5234 & 0.5399 & 0.5569 \\
 & LPIPS $\downarrow$ & \best{0.3043} & 0.4036 & 0.4967 & 0.3773 & 0.3158 & 0.3564 & 0.3311 & \second{0.3129} \\
 & NIQE $\downarrow$ & 3.0804 & 3.0017 & 5.2237 & 4.2282 & 3.5157 & 3.4153 & \second{2.9504} & \best{2.9368} \\
 & MUSIQ $\uparrow$ & 59.12 & 60.02 & 37.95 & 50.44 & 57.65 & \best{63.77} & 56.01 & \second{60.42} \\
 & CLIP-IQA $\uparrow$ & 0.3236 & \second{0.3698} & 0.2132 & 0.2904 & 0.3041 & \best{0.3914} & 0.3160 & 0.3396 \\
 & DOVER $\uparrow$ & 0.3293 & 0.2801 & 0.2068 & 0.3337 & 0.3679 & \second{0.3893} & 0.3477 & \best{0.3948} \\
\midrule

\multirow{7}{*}{\textbf{UDM10}}
 & PSNR $\uparrow$ & \best{26.65} & 24.68 & 25.49 & \second{26.39} & 25.99 & 25.41 & 24.41 & 26.02 \\
 & SSIM $\uparrow$ & 0.7483 & 0.6991 & 0.7237 & \second{0.7504} & 0.7313 & 0.7213 & 0.7008 & \best{0.7626} \\
 & LPIPS $\downarrow$ & 0.2670 & 0.3070 & 0.3677 & 0.2341 & \second{0.2150} & 0.2640 & 0.2528 & \best{0.1987} \\
 & NIQE $\downarrow$ & 4.4199 & 4.6452 & 7.1730 & 4.7713 & 4.6516 & \second{4.2548} & \best{4.0032} & 4.5547 \\
 & MUSIQ $\uparrow$ & 56.98 & 57.57 & 31.50 & 60.30 & 58.49 & \second{64.55} & 64.24 & \best{64.59} \\
 & CLIP-IQA $\uparrow$ & 0.3705 & 0.3746 & 0.2225 & 0.4717 & 0.4000 & \second{0.4986} & \best{0.4987} & 0.4862 \\
 & DOVER $\uparrow$ & 0.4646 & 0.4232 & 0.2455 & 0.4996 & 0.5013 & 0.4744 & \second{0.5507} & \best{0.5587} \\
\midrule

\multirow{7}{*}{\textbf{YouHQ40}}
 & PSNR $\uparrow$ & \second{24.17} & 22.98 & 23.46 & \best{24.29} & 23.66 & 23.08 & 22.68 & 23.56 \\
 & SSIM $\uparrow$ & 0.6414 & 0.6109 & 0.6429 & \best{0.6662} & \second{0.6555} & 0.6151 & 0.6026 & 0.6434 \\
 & LPIPS $\downarrow$ & 0.3408 & 0.3508 & 0.4545 & 0.2920 & \second{0.2726} & 0.2912 & 0.2742 & \best{0.2431} \\
 & NIQE $\downarrow$ & 3.6506 & 3.8166 & 6.9886 & 4.3244 & 4.1732 & 3.2815 & \best{3.1768} & \second{3.2469} \\
 & MUSIQ $\uparrow$ & 59.63 & 56.09 & 32.91 & 60.63 & 57.69 & 62.01 & \second{65.79} & \best{66.34} \\
 & CLIP-IQA $\uparrow$ & 0.3996 & 0.4024 & 0.2651 & 0.4500 & 0.3970 & 0.5128 & \second{0.5333} & \best{0.5456} \\
 & DOVER $\uparrow$ & 0.6149 & 0.5763 & 0.4308 & 0.6649 & 0.6709 & 0.6877 & \second{0.6916} & \best{0.7224} \\
\midrule

\multirow{4}{*}{\textbf{VideoLQ}}
 & NIQE $\downarrow$ & 4.3616 & 4.9571 & 5.6624 & 5.0218 & 4.7326 & 4.4435 & \best{3.9488} & \second{4.1491} \\
 & MUSIQ $\uparrow$ & 49.22 & 44.13 & 35.29 & 44.91 & 41.68 & 50.46 & \second{50.73} & \best{52.82} \\
 & CLIP-IQA $\uparrow$ & 0.3263 & 0.2837 & 0.2454 & 0.2942 & 0.2428 & \second{0.3583} & \best{0.3623} & 0.3552 \\
 & DOVER $\uparrow$ & 0.4621 & 0.4204 & 0.4135 & 0.4972 & 0.4455 & 0.4957 & \second{0.5339} & \best{0.5567} \\

\bottomrule
\end{tabular}
}
\par
\endgroup
\end{table}

\vspace{-2pt}
\section{Experiments}
\label{sec:exp}
\vspace{-2pt}
\subsection{Experimental Setup}
\vspace{-2pt}
\paragraph{Implementation details.}
ReCaVSR is initialized from the pretrained Wan2.2-TI2V-5B
and adapted using LoRA with rank 512,
together with trainable LR and SR latent projections.
We train on a self-collected dataset comprising about 0.5M videos and 1M images,
with paired LR--HQ samples synthesized using the degradation pipeline
of RealBasicVSR~\citep{chan2022investigating}.
We optimize the generator with AdamW on 85-frame HQ video clips and
single HQ images at $704\times1280$.
The LR-conditioned FlashDecoder~\citep{kang2026flashdecoder} is trained
separately on 17-frame HQ clips and images at $896\times1344$.
\vspace{-2pt}
\paragraph{Evaluation datasets and protocol.}
We evaluate on three synthetic benchmarks---UDM10, YouHQ40,
and REDS30---and the real-world benchmark VideoLQ\@.
Synthetic LR inputs are generated using the same degradation pipeline as training.
To assess long-video performance, we additionally evaluate on LongVSR60,
comprising 30 real-world and 30 AI-generated single-shot videos
with approximately 1000 frames each and no paired HQ references.
On the synthetic benchmarks, we report PSNR~\citep{huynh2008scope}
and SSIM~\citep{wang2004image} for reconstruction fidelity
and LPIPS~\citep{zhang2018unreasonable} for perceptual quality.
We additionally report NIQE~\citep{mittal2012making},
MUSIQ~\citep{ke2021musiq}, CLIP-IQA~\citep{wang2023exploring},
and DOVER~\citep{wu2023dover} as no-reference quality measures.
For LongVSR60, we further report subject consistency (SC),
background consistency (BC), and motion smoothness (MS).
Additional training details and baseline configurations are provided in the appendix.

\subsection{Comparison with Existing Methods}
\begin{figure}[!t]
\centering
\includegraphics[width=\linewidth]{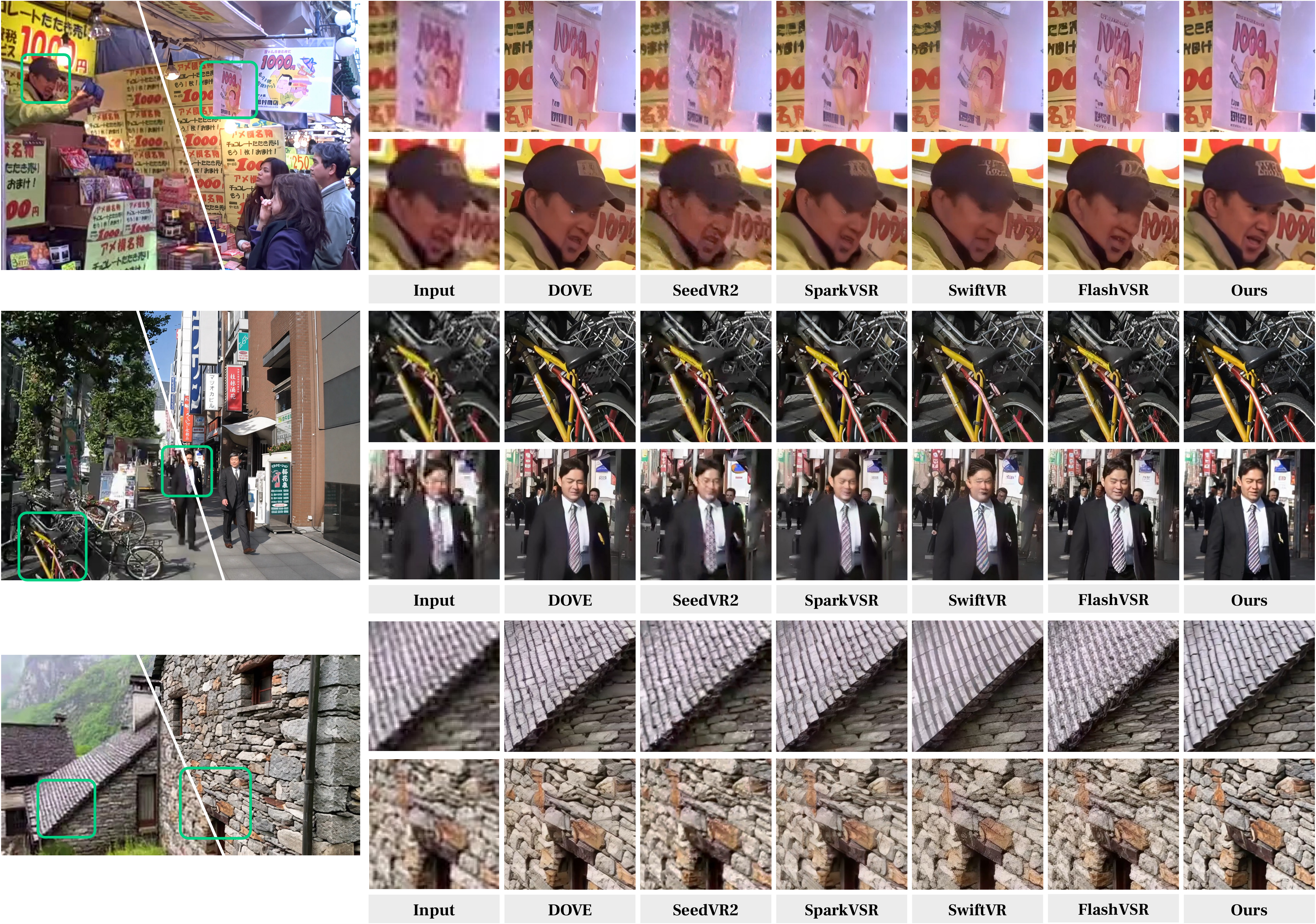}
\setlength{\abovecaptionskip}{0pt}
\caption{Qualitative comparisons on challenging synthetic and real-world VSR examples.
ReCaVSR restores sharper structures, more natural textures,
and cleaner face, text, and logo details.}
\label{fig:qualitative}
\vspace{-10pt}
\end{figure}

\vspace{-2pt}
\paragraph{Quantitative comparisons.}
We compare ReCaVSR with seven representative VSR methods:
RealViformer~\citep{zhang2024realviformer},
Upscale-A-Video (UAV)~\citep{zhou2024upscale}, STAR~\citep{xie2025star},
DOVE~\citep{chen2025dove}, SeedVR2-3B~\citep{wang2025seedvr2},
SwiftVR~\citep{yan2026swiftvr}, and FlashVSR-Tiny~\citep{zhuang2025flashvsr}.
Table~\ref{tab:main_results} reports results on three synthetic benchmarks and VideoLQ,
where ReCaVSR achieves the highest DOVER scores across all four datasets,
indicating consistent gains in overall video quality on both synthetic and real-world inputs.
On the synthetic benchmarks, ReCaVSR obtains the lowest LPIPS on UDM10 and YouHQ40
and the second-lowest on REDS30.
Compared with SeedVR2-3B, it reduces LPIPS from 0.2150 to 0.1987 on UDM10
and from 0.2726 to 0.2431 on YouHQ40.
ReCaVSR also outperforms SwiftVR and FlashVSR-Tiny in PSNR, SSIM, and LPIPS
across all three synthetic benchmarks,
demonstrating improvements in both reconstruction fidelity and perceptual similarity.
On the real-world VideoLQ benchmark, ReCaVSR achieves the highest MUSIQ and DOVER scores
of 52.82 and 0.5567, respectively.
\vspace{-2pt}
\paragraph{Qualitative comparisons.}
Figure~\ref{fig:qualitative} presents visual comparisons on challenging examples
containing text, faces, thin structures, and repeated textures.
ReCaVSR produces well-defined character contours and facial features,
while several competing methods exhibit blurred details or distorted strokes.
In the bicycle region, it resolves thin frame edges and spokes within a cluttered background.
The roof example further highlights its reconstruction of repeated structures:
ReCaVSR produces distinct tile boundaries and a regular arrangement,
whereas SwiftVR yields overly smooth bands and FlashVSR introduces grainy surface textures.
In the stone-wall region, ReCaVSR reconstructs clear stone boundaries and joints
with less distracting texture.
These examples illustrate its ability to generate fine details
while maintaining coherent local structures.

\begin{table}[H]
\begingroup
\centering
\caption{Quantitative comparison on LongVSR60 (Real + AIGC).
The best and second-best results are marked in \best{red} and \second{blue}, respectively.}
\label{tab:longvsr60_results}
\renewcommand{\arraystretch}{1.1}
\resizebox{0.9\textwidth}{!}{\begin{tabular}{@{}lrrrrrrr@{}}
\toprule
Method & NIQE $\downarrow$ & MUSIQ $\uparrow$ & CLIP-IQA $\uparrow$ & DOVER $\uparrow$ & SC $\uparrow$ & BC $\uparrow$ & MS $\uparrow$ \\
\midrule
RealViformer & 5.0770 & 46.17 & 0.3819 & 0.5053 & \best{0.8846} & \best{0.9198} & \best{0.9848} \\
Stream-DiffVSR & 4.0491 & 54.52 & 0.4694 & 0.5704 & 0.8830 & 0.9161 & 0.9813 \\
SwiftVR & 4.2822 & 53.89 & 0.4640 & 0.5512 & 0.8831 & \second{0.9177} & 0.9824 \\
FlashVSR & \best{3.8160} & \second{55.15} & \second{0.4747} & \second{0.5895} & 0.8829 & 0.9144 & 0.9802 \\
Ours & \second{3.9921} & \best{57.89} & \best{0.4932} & \best{0.6075} & \second{0.8834} & 0.9133 & \second{0.9828} \\
\bottomrule
\end{tabular}
}
\par
\endgroup
\end{table}

\subsection{Long-Video Evaluation and Efficiency}
\vspace{-2pt}
\paragraph{Long-video performance.}
We evaluate ReCaVSR on LongVSR60 to assess its performance
on sequences of approximately 1,000 frames.
As shown in Table~\ref{tab:longvsr60_results}, ReCaVSR achieves the highest
MUSIQ, CLIP-IQA, and DOVER scores of 57.89, 0.4932, and 0.6075, respectively,
outperforming FlashVSR by 2.74, 0.0185, and 0.0180.
We report SC, BC, and MS as supplementary temporal indicators,
as they can favor overly smooth outputs.
Their scores alone cannot establish the temporal consistency of fine textures.
Overall, the results demonstrate strong perceptual quality on long videos
containing real-world and AI-generated content.

\newsavebox{\cacheAblationCaption}
\newsavebox{\queryAblationCaption}
\newlength{\ablationCaptionHeight}

\vspace{-2pt}
\paragraph{Inference efficiency.}
We benchmark GPU inference on 201 real input frames at
$1080{\times}1920$ output resolution using a single NVIDIA A100-80GB GPU.
After warm-up, we measure throughput and first-output latency with CUDA events
and report peak allocated GPU memory.
First-output latency denotes the cumulative GPU model time until the first complete
RGB output block is available: 21 frames for FlashVSR and one frame for Stream-DiffVSR.
We use the Tiny variant of FlashVSR.
As shown in Table~\ref{tab:efficiency_comparison}, ReCaVSR achieves the highest throughput
of 21.20 FPS and the lowest peak memory usage of 15.16 GB among the compared methods.
Compared with FlashVSR, it delivers $2.72\times$ the throughput
while reducing peak memory usage by 38.0\%.
ReCaVSR also produces its first output in 0.982\,s,
compared with 2.830\,s for FlashVSR.

\begin{table}[H]
\begingroup
\vspace{-10pt}
\centering
\caption{GPU inference efficiency at $1080{\times}1920$ output resolution
on 201 real input frames using a single NVIDIA A100-80GB GPU.}
\label{tab:efficiency_comparison}
\renewcommand{\arraystretch}{1.1}  \resizebox{\textwidth}{!}{
\begin{tabular}{@{}lcccccc@{}}
\toprule
Metric & DOVE & SeedVR2-3B & SparkVSR & Stream-DiffVSR & FlashVSR & ReCaVSR \\
\midrule
First-output latency (s) $\downarrow$ & 356.801 & 255.199 & 356.482 & \textbf{0.705} & 2.830 & \underline{0.982} \\
FPS $\uparrow$ & 0.563 & 0.788 & 0.564 & 0.849 & \underline{7.799} & \textbf{21.202} \\
Peak Mem. (GB) $\downarrow$ & 41.079 & 73.889 & 41.296 & 25.923 & \underline{24.447} & \textbf{15.159} \\
\bottomrule
\end{tabular}
}
\par
\endgroup
\end{table}

\begin{table}[!t]
\centering
\sbox{\cacheAblationCaption}{\begin{minipage}{0.55\textwidth}
\caption{Recycling and cache allocation on UDM10.
Uniform variants use rolling windows of $R$ or $3R$ latent positions per layer.
FLOPs count historical attention.
$\dagger$: without SR latent recycling.}
\label{tab:ablation_recycle_cache}
\end{minipage}}
\sbox{\queryAblationCaption}{\begin{minipage}{0.42\textwidth}
\caption{Query-scope ablation on UDM10 with a matched query budget.
The best and second-best results are bold and underlined, respectively.}
\label{tab:ablation_msq}
\label{tab:ablation_components}
\end{minipage}}
\setlength{\ablationCaptionHeight}{\dimexpr\ht\cacheAblationCaption+\dp\cacheAblationCaption\relax}
\ifdim\dimexpr\ht\queryAblationCaption+\dp\queryAblationCaption\relax>\ablationCaptionHeight
\setlength{\ablationCaptionHeight}{\dimexpr\ht\queryAblationCaption+\dp\queryAblationCaption\relax}
\fi
\begin{minipage}[c]{0.55\textwidth}
\centering
\begin{minipage}[c][\ablationCaptionHeight][c]{\linewidth}
\usebox{\cacheAblationCaption}
\end{minipage}\par\vspace{6pt}
\scriptsize
\setlength{\tabcolsep}{2pt}
\renewcommand{\arraystretch}{1.2}
\begin{tabular*}{\linewidth}{@{\extracolsep{\fill}}lcccc@{}}
\toprule
Variant & DOVER $\uparrow$
& MS $\uparrow$
& KV (GB) $\downarrow$
& FLOPs $\downarrow$ \\
\midrule

Uniform ($3R$) & \textbf{0.5601} & \textbf{0.9801} & 6.44 & $1.00\times$ \\
Uniform ($R$) & 0.5261 & 0.9722 & \textbf{2.15} & $\mathbf{0.27}\times$ \\
Ours$^{\dagger}$ & 0.5130 & 0.9748 & \underline{2.17} & $\underline{0.28\times}$ \\
Ours & \underline{0.5587} & \underline{0.9785} & \underline{2.17} & $\underline{0.28\times}$ \\
\bottomrule
\end{tabular*}
\end{minipage}\hfill
\begin{minipage}[c]{0.42\textwidth}
\centering
\begin{minipage}[c][\ablationCaptionHeight][c]{\linewidth}
\usebox{\queryAblationCaption}
\end{minipage}\par\vspace{6pt}
\scriptsize
\setlength{\tabcolsep}{1.5pt}
\renewcommand{\arraystretch}{1.2}
\begin{tabular*}{\linewidth}{@{\extracolsep{\fill}}lccc@{}}
\toprule
Query scopes & NIQE $\downarrow$ & MUSIQ $\uparrow$
& DOVER $\uparrow$ \\
\midrule
Global only & 4.8241 & 61.80 & 0.5402 \\
Global + Spatial & \underline{4.6332} & \underline{64.10} & 0.5480 \\
Global + Temporal & 4.7238 & 63.20 & \underline{0.5520} \\
Ours & \textbf{4.5547} & \textbf{64.59} & \textbf{0.5587} \\
\bottomrule
\end{tabular*}
\end{minipage}
\end{table}

\subsection{Ablation Studies}
We examine three design choices that shape ReCaVSR's quality and efficiency.
The first study isolates the contributions of recycled SR latents and
layer-wise KV allocation across cache budgets. The second tests spatial and
temporal adversarial feedback, while the third evaluates how LR conditioning
changes reconstruction quality and decoding cost.

\paragraph{Recycled SR latents and layer-wise cache allocation.}
Table~\ref{tab:ablation_recycle_cache} compares uniform and learned cache allocations,
with Uniform ($R$) using a comparable exported cache budget to Ours.
Ours$^{\dagger}$ disables latent recycling while retaining the same exported route.
Removing recycling reduces DOVER from 0.5587 to 0.5130,
indicating the contribution of the recurrent SR condition.
At a comparable cache budget, learned allocation improves DOVER from 0.5261 to 0.5587
over Uniform ($R$).
Compared with Uniform ($3R$), Ours reduces KV memory from 6.44 to 2.17 GB
and historical-attention FLOPs to $0.28\times$, with a DOVER decrease of only 0.0014.
MS is reported as a supplementary temporal indicator. At the matched
cache budget, Ours raises MS from 0.9722 to 0.9785 relative to
Uniform ($R$), approaching the 0.9801 score of Uniform ($3R$).

\paragraph{Multi-scope adversarial supervision.}
Table~\ref{tab:ablation_msq} compares global-only queries,
global queries combined with either spatial-window or temporal-tube queries,
and the complete MSQ design under a matched total query budget,
with the discriminator backbone and training settings fixed.
Both partial variants improve all three metrics over the global-only control.
Among these two variants, spatial-window queries yield lower NIQE and higher MUSIQ,
while temporal-tube queries achieve higher DOVER.
Combining all three scopes achieves the best NIQE, MUSIQ, and DOVER scores
of 4.5547, 64.59, and 0.5587, respectively,
supporting the complementary contributions of spatial and temporal query scopes
to perceptual quality. Relative to global-only queries, the complete design
reduces NIQE by 0.2694 and increases MUSIQ by 2.79 and DOVER by 0.0185.

\vspace{6pt}
\noindent
\begin{minipage}[c]{0.55\textwidth}
\paragraph{Decoder quality and efficiency.}
Table~\ref{tab:ablation_decoder} compares RGB reconstruction quality on 50 clips
of 37 frames at 720p, using each decoder's matching encoder.
Decode-only throughput and peak allocated GPU memory are measured on 201 frames
at 1080p using an A100-80GB GPU.
Our LR-conditioned FlashDecoder achieves a PSNR of 36.2525 and a throughput of 91.71 FPS,
with a peak memory usage of 1.33 GB.
Relative to Wan2.2 VAE, it delivers $24.33\times$ the decoding throughput
while reducing peak memory by 94.8\%, supporting efficient latent decoding
in our streaming VSR pipeline. LR conditioning improves reconstruction PSNR
from 34.4523 to 36.2525 (1.80\,dB), while decode-only throughput remains
similar at 93.20 versus 91.71 FPS.
\end{minipage}\hfill
\begin{minipage}[c]{0.42\textwidth}
\centering
\setlength{\abovecaptionskip}{0pt}
\captionof{table}{Decoder quality and efficiency.
$\ddagger$: without LR conditioning.
Best and second-best values are bold and underlined.}
\label{tab:ablation_decoder}
\footnotesize
\setlength{\tabcolsep}{2pt}
\renewcommand{\arraystretch}{1.1}
\begin{tabular*}{\linewidth}{@{\extracolsep{\fill}}lccc@{}}
\toprule
Decoder & PSNR $\uparrow$ & FPS $\uparrow$
& Mem. (GB) \\
\midrule
Wan2.1 VAE & \underline{38.5332} & 4.51 & 22.13 \\
Wan2.2 VAE & \textbf{39.1866} & 3.77 & 25.80 \\
TCDecoder & 36.8953 & 33.06 & 5.56 \\
SwiftVR ReAE & 30.9792 & \textbf{141.35} & 18.11 \\
\midrule
Ours$^{\ddagger}$ & 34.4523 & \underline{93.20} & \textbf{1.28} \\
Ours & 36.2525 & 91.71 & \underline{1.33} \\
\bottomrule
\end{tabular*}
\end{minipage}
\par\medskip

Additional visual analysis and limitations
are provided in Appendices~\ref{app:qualitative} and~\ref{app:limitations}.

\section{Conclusion}
\label{sec:conclusion}

ReCaVSR shows that one-step streaming video super-resolution can use temporal
history selectively: recycled SR latents propagate local context, while
layer-wise routing retains historical K/V only where needed. Sequential
self-rollout aligns training with causal inference, while multi-scope
adversarial supervision and LR-conditioned decoding support fine detail and
temporal stability. Ablations confirm that latent recycling and learned cache
allocation each contribute to perceptual quality. ReCaVSR achieves the highest
DOVER on four benchmarks and the highest quality, detail, and temporal-stability
ratings in our 35-video human study. At 1080p on a single A100-80GB, it runs at 21.20 FPS,
$2.72\times$ faster than FlashVSR Tiny while using 38.0\% less peak GPU
memory. Its first complete RGB output takes 0.982\,s of GPU model time,
complementing the throughput gain with low startup latency in causal
streaming. These results support selective temporal memory as an effective
design for high-quality streaming VSR.

\appendix
\clearpage

\begin{center}
    \Large\textbf{{Appendix}}\\
    \vspace{8mm}
\end{center}

\section{Additional Method Details}
\label{app:method_details}
\label{app:streaming_inference}
\setcounter{figure}{0}
\setcounter{table}{0}
\renewcommand{\thefigure}{\thesection.\arabic{figure}}
\renewcommand{\thetable}{\thesection.\arabic{table}}
\renewcommand{\theHfigure}{appendix.\thesection.\arabic{figure}}
\renewcommand{\theHtable}{appendix.\thesection.\arabic{table}}

\begin{figure}[htbp]
    \centering
    \includegraphics[width=\linewidth]{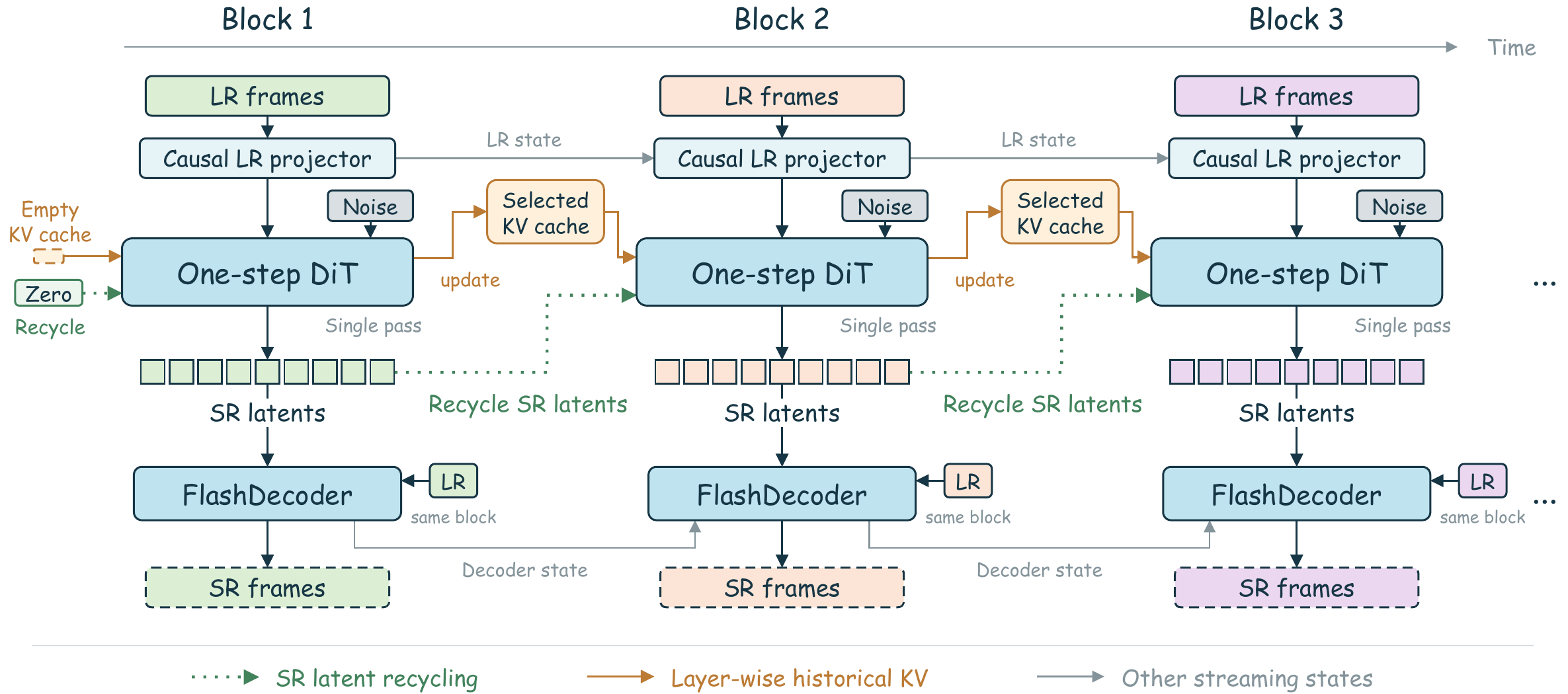}
    \caption{\textbf{Blockwise streaming inference with ReCaVSR.}
    Each block uses one DiT forward pass to generate SR latents, which are
    decoded with LR observations from the same block.
    Green dotted paths recycle the generated SR latents; orange paths carry
    the historical KV selected for each layer; gray paths preserve the LR
    projector and decoder states.
    The first block starts with zero recycled input and an empty DiT KV cache.}
    \label{fig:app_streaming_inference}
\end{figure}

This section expands the input alignment, attention supports, and state updates
underlying Sec.~\ref{sec:method}.
Figure~\ref{fig:app_streaming_inference} summarizes the streaming execution,
combining latent recycling, layer-wise cache routing, and streaming decoding.

An incomplete final block repeats its last LR frame to obtain the required
input shape; its output is trimmed to the original valid frame count.
No future block is needed for any preceding output block.

\subsection{Input Alignment and Recycled Latent Conditioning}
\label{app:streaming_formulation}

\paragraph{Input embedding and conditioning.}
Let $r_k$ denote the recurrent SR condition for block $k$.
Noise and recycled latents pass through the same patch embedding $P$;
the recycled features then pass through an additional linear projection $S$.
The causal LR projector $C_{LR}$ provides the other additive condition:
\begin{equation}
    h_k^{\mathrm{in}}=
    P(\epsilon_k)+C_{LR}(x_k^{LR})+S\!\left(P(r_k)\right).
    \label{eq:app_conditioning}
\end{equation}
The convolutional embedding $P$ gives noise and recycled latents the same
token layout, while $S$ is a token-wise linear projection.
The LR features are aligned to this layout before addition.

\paragraph{Recurrent initialization and update.}
The first block uses a zero recurrent condition matching its latent shape.
For each subsequent block, $r_k=\widehat z_{k-1}^{SR}$ contains all SR latents
predicted for the preceding block. The complete block is embedded and
projected as the additive recurrent condition for the current prediction.

\paragraph{Causal LR projection.}
We follow the causal LR projection design of FlashVSR~\citep{zhuang2025flashvsr}.
LR frames are bilinearly upsampled to the target resolution and normalized
before projection. Pixel unshuffle converts local spatial neighborhoods into
channels; causal spatiotemporal convolutions with RMS normalization and SiLU
activations then align the temporal and spatial grid with $P$.
A final linear projection matches the DiT feature width.
Each convolution retains its own input history across blocks, so incremental
projection continues the same causal computation without restarting at block
boundaries.

\subsection{Layer-Wise Cache Routing and State Maintenance}
\label{app:routed_attention}

During streaming inference, each DiT layer attends to the current block
together with the historical K/V selected by its assigned action.
The first block starts with empty historical caches.
For later blocks, recent and anchor histories are combined, with overlapping
positions included once. A layer assigned $\emptyset$ uses current-block
tokens for self-attention and retains no historical K/V.

After generating a block, each layer updates its history using the K/V
produced in that same forward pass. The recent window moves forward to retain
the latest latent positions. The anchor history incorporates newly eligible
positions and removes its oldest entries when its capacity is reached.
These updated histories are then used by the next block.
Fixed physical slots map absolute latent indices to cached K/V; empty slots
are masked until populated. Temporal RoPE uses the advancing time origin
when cached states are read, without recomputing their K/V.
Each layer's selection rule remains fixed throughout the video, while the
retained positions advance with the stream. This maintains a bounded amount
of historical context as the video grows. The spatial attention window is
independent of this temporal cache policy.

\subsection{Multi-Scope Query Discriminator}
\label{app:msq_details}

\paragraph{Query support construction.}
Let $F_{t,u,v}$ denote a projected feature token at temporal position $t$ and
spatial position $(u,v)$.
For a query $q$, let $\Omega_q$ be its spatial window and $\mathcal{I}_q$
its temporal interval; $t_q$ denotes the selected latent position for a spatial query.
The three query groups use the following supports:
\begin{equation}
    \begin{aligned}
        \mathcal{S}_{\mathrm{global},q}&=\{(t,u,v):\text{all clip positions}\},\\
        \mathcal{S}_{\mathrm{spatial},q}&=\{(t_q,u,v):(u,v)\in\Omega_q\},\\
        \mathcal{S}_{\mathrm{temporal},q}&=\{(t,u,v):t\in\mathcal{I}_q,
            (u,v)\in\Omega_q\}.
    \end{aligned}
    \label{eq:app_msq_supports}
\end{equation}
These sets define the keys and values visible to each query in cross-attention.
The temporal-tube support uses the same spatial window throughout its interval.
Its temporal extent belongs to the discriminator's clip-level supervision and
does not impose a causal mask on feature extraction or change the generator's
streaming visibility.

\subsection{Inference at Arbitrary Spatial Resolutions}
\label{app:arbitrary_resolution}

ReCaVSR accommodates different spatial resolutions by constructing the
inference grid from the requested output dimensions.
We bilinearly upsample the LR frames to the target size and pad their height
and width to multiples of 32, aligning the LR conditioning features with
the latent grid and DiT patch embedding.
Spatial rotary embeddings are constructed for this grid and shared by
current and cached tokens throughout the video.
As the grid grows, we bound each query's spatial attention support using
the locality-constrained window design of FlashVSR~\citep{zhuang2025flashvsr}.
The window is applied across the current block and the historical positions
selected by each layer's cache action, coupling local spatial processing
with layer-specific temporal context.
At image boundaries, the window shifts inward to preserve its extent;
dimensions smaller than the configured window are covered in full.
The generated latents and aligned LR observations then pass through the
streaming decoder, whose RGB outputs are cropped to the requested dimensions.

\section{Training Details}
\label{app:training_details}
\setcounter{table}{0}

\subsection{Model Configuration}
\label{app:model_configuration}

Table~\ref{tab:app_model_configuration} lists the architecture settings.
LoRA updates the attention projections and feed-forward linear layers.
Anchor positions follow $5+6m$ (zero-based), retaining the two latest
eligible positions; combined cache actions use the union of recent and
anchor histories described in Appendix~\ref{app:routed_attention}.

\begin{table}[H]
\centering
\caption{Architecture configurations of ReCaVSR.}
\label{tab:app_model_configuration}
\begingroup
\small
\setlength{\tabcolsep}{0pt}
\renewcommand{\arraystretch}{1.08}
\begin{tabular*}{\textwidth}{@{\extracolsep{\fill}}lr@{\hspace{22pt}}lr@{}}
\toprule
\multicolumn{2}{@{}l}{\textbf{Generator}} &
\multicolumn{2}{l@{}}{\textbf{LR-conditioned decoder}} \\
\cmidrule(r{10pt}){1-2}\cmidrule(l{10pt}){3-4}
Backbone & Wan2.2-TI2V-5B & Architecture & FlashDecoder \\
DiT layers & 30 & Backbone layers & 12 \\
Hidden dimension & 3072 & Refinement layers & 2 \\
Latent channels & 48 & Hidden dimension & 512 \\
Patch size $(T,H,W)$ & $1\times2\times2$ & Latent projection & $48\rightarrow512$ \\
LoRA rank & 512 & LR stem & $768\rightarrow48$ \\
Recycle projection & $3072\rightarrow3072$ & Grouped LR projection & $192\rightarrow512$ \\
LR output dimension & 3072 & Frame-aligned LR projection & $48\rightarrow512$ \\
Spatial window & $22\times40$ & Parameters & 57.17M \\
\midrule
\multicolumn{2}{@{}l}{\textbf{Cache router}} &
\multicolumn{2}{l@{}}{\textbf{MSQ discriminator}} \\
\cmidrule(r{10pt}){1-2}\cmidrule(l{10pt}){3-4}
Layer embedding dimension & 64 & Feature layers (1-based) & 8, 15, 22 \\
MLP dimensions & $64\rightarrow128\rightarrow7$ & Projected feature dimension & 512 \\
Activation & SiLU & Queries per scope per depth & 4 \\
Number of actions & 7 & Total queries & 36 \\
Recent window sizes & $\{1,2,4\}$ & Global support & Full clip \\
Anchor interval & 6 & Spatial support $(T,H,W)$ & $1\times8\times8$ \\
First anchor index (0-based) & 5 & Temporal support $(T,H,W)$ & $4\times8\times8$ \\
Anchor capacity & 2 & Scope weights & $1/3,\;1/3,\;1/3$ \\
\bottomrule
\end{tabular*}
\endgroup
\end{table}

\subsection{Stage 1: VSR Adaptation and Route Learning}
\label{app:stage1_training}

\paragraph{VSR adaptation.}
We initialize the generator from Wan2.2-TI2V-5B and freeze its pretrained
weights, updating the LoRA adapters and the LR and recycled-latent projections.
Training uses 85-frame LR--HQ clips and images at $704\times1280$,
with batch size $32$ and an image-batch fraction of $0.25$.
Both phases use AdamW with learning rate $2\times10^{-5}$,
$(\beta_1,\beta_2)=(0.9,0.99)$, and weight decay $10^{-4}$.
The frozen Wan encoder supplies HQ latents for training.
For each video block, the recurrent condition contains the complete clean HQ
latent block immediately preceding it; the first block and image samples
use zero recurrent conditions.
For HQ latent $z$, $\epsilon\sim\mathcal{N}(0,I)$, and logit-normal
$t=(1+\exp(-u))^{-1}$ with $u\sim\mathcal{N}(0,1)$, the flow-matching
input and velocity target are $z_t=(1-t)z+t\epsilon$ and
$v^*=\epsilon-z$. With $\mathrm{cond}$ denoting LR and recurrent
conditions, Stage~1 uses
\begin{equation}
\mathcal{L}_{\mathrm{Stage1}}
 =\bigl\|G_{\theta}(z_t,t,\mathrm{cond})-v^*\bigr\|_2^2
 +\lambda_{\mathrm{budget}}(\bar c-c_{\mathrm{target}})^2
 +\frac{\lambda_{\mathrm{sharp}}}{L}\sum_{l=1}^{L}H(\pi_l).
\label{eq:app_stage1_loss}
\end{equation}
Here $\bar c$ is the expected per-layer capacity defined in
Eq.~\ref{eq:stage1_objective}. During the first 70K adaptation updates,
each layer uses $W_6$; the router is frozen and the regularization terms
are disabled.

\paragraph{Route learning.}
For 30K further updates, we jointly optimize the router, its layer
embeddings, and the same generator parameters while retaining clean HQ
recurrent conditions.
We set $c_{\mathrm{target}}=2.0$,
$\lambda_{\mathrm{budget}}=\lambda_{\mathrm{sharp}}=0.5$, and anneal
$\tau$ from $2$ to $0.3$ during route learning.
At its completion, we export the highest-probability action for each layer
from the final router checkpoint and initialize Stage~2 with the corresponding
generator checkpoint. The exported schedule remains fixed throughout Stage~2.

\subsection{Stage 2: One-Step Adversarial Training}
\label{app:stage2_training}

Stage~2 runs for 10K updates on 85-frame LR--HQ clips and images at
$704\times1280$, with video and image batch sizes of $16$ and $64$
and an image-batch fraction of $0.2$. Each video yields a six-latent
prefix and eight two-latent blocks, generated sequentially in one step
at $t=1000$. The prefix starts with zero recurrent input; later blocks
recycle the preceding SR latents. Recycled latents and historical KV
states are detached across blocks, while all blocks receive latent and
adversarial supervision. Real and generated clips share Gaussian noise
$\xi$ and $\sigma_D=\max(t_D,1)/1000$ with
$t_D\sim\operatorname{Unif}\{0,\ldots,50\}$.

For generated latents $\widehat z$ and HQ latents $z$, the Stage~2 losses are
\begin{equation}
    \begin{gathered}
        \mathcal{L}_{\mathrm{stage2}}
        =\|\widehat z-z\|_2^2
        +0.1\mathcal{L}_{\mathrm{adv}}^{G}
        +\mathcal{L}_{\mathrm{RGB}}
        +2\mathcal{L}_{\mathrm{perc}},\\
        \mathcal{L}_{D}
        =\mathcal{L}_{\mathrm{adv}}^{D}
        +1000\mathcal{L}_{\mathrm{fR1}}.
    \end{gathered}
    \label{eq:app_stage2_losses}
\end{equation}
The adversarial terms are defined in Eq.~\ref{eq:rsgan_objective}.
RGB MSE and LPIPS supervise all 85 frames decoded from the complete latent
sequence through the frozen Wan decoder; image samples use their single
available frame.

Generator and discriminator both use AdamW with learning rate $10^{-5}$
and weight decay $10^{-4}$; their $\beta$ values are $(0.5,0.99)$ and
$(0,0.99)$, respectively. The discriminator starts at step 0 with a
20-update learning-rate warm-up. Generator updates and its EMA
(decay $0.999$) start at step 20.

\subsection{LR-Conditioned FlashDecoder Training}
\label{app:decoder_training}

The decoder is trained independently of Stage~2. Given HQ RGB $x^{HQ}$
and paired LR input $x^{LR}$, let $z^{HQ}=E_{\mathrm{Wan}}(x^{HQ})$
be the latent from the frozen Wan encoder and
$\widehat{x}=D_{\phi}(z^{HQ},x^{LR})$ the decoded prediction.
The reconstruction objective is
\begin{equation}
\mathcal{L}_{\mathrm{dec}}
=\bigl\|\widehat{x}-x^{HQ}\bigr\|_1
 +\lambda_{\mathrm{dec,perc}}\operatorname{LPIPS}(\widehat{x},x^{HQ}).
\label{eq:app_decoder_loss}
\end{equation}
We train the decoder on 17-frame clips and images at $896\times1344$,
with video and image batch sizes of $32$ and $128$ and an image-batch
fraction of $0.2$. AdamW runs for up to 500K updates with learning rate
$5\times10^{-5}$,
$(\beta_1,\beta_2)=(0.9,0.999)$, weight decay $0.01$, and EMA decay
$0.9995$ and $\lambda_{\mathrm{dec,perc}}=0.4$.

\section{Additional Results and Analysis}
\label{app:additional_results}

\subsection{Additional Quality Evaluation}
\label{app:qualitative}

\begin{figure}[H]
\centering
\renewcommand{\thefigure}{C.\arabic{figure}}
\includegraphics[width=\textwidth]{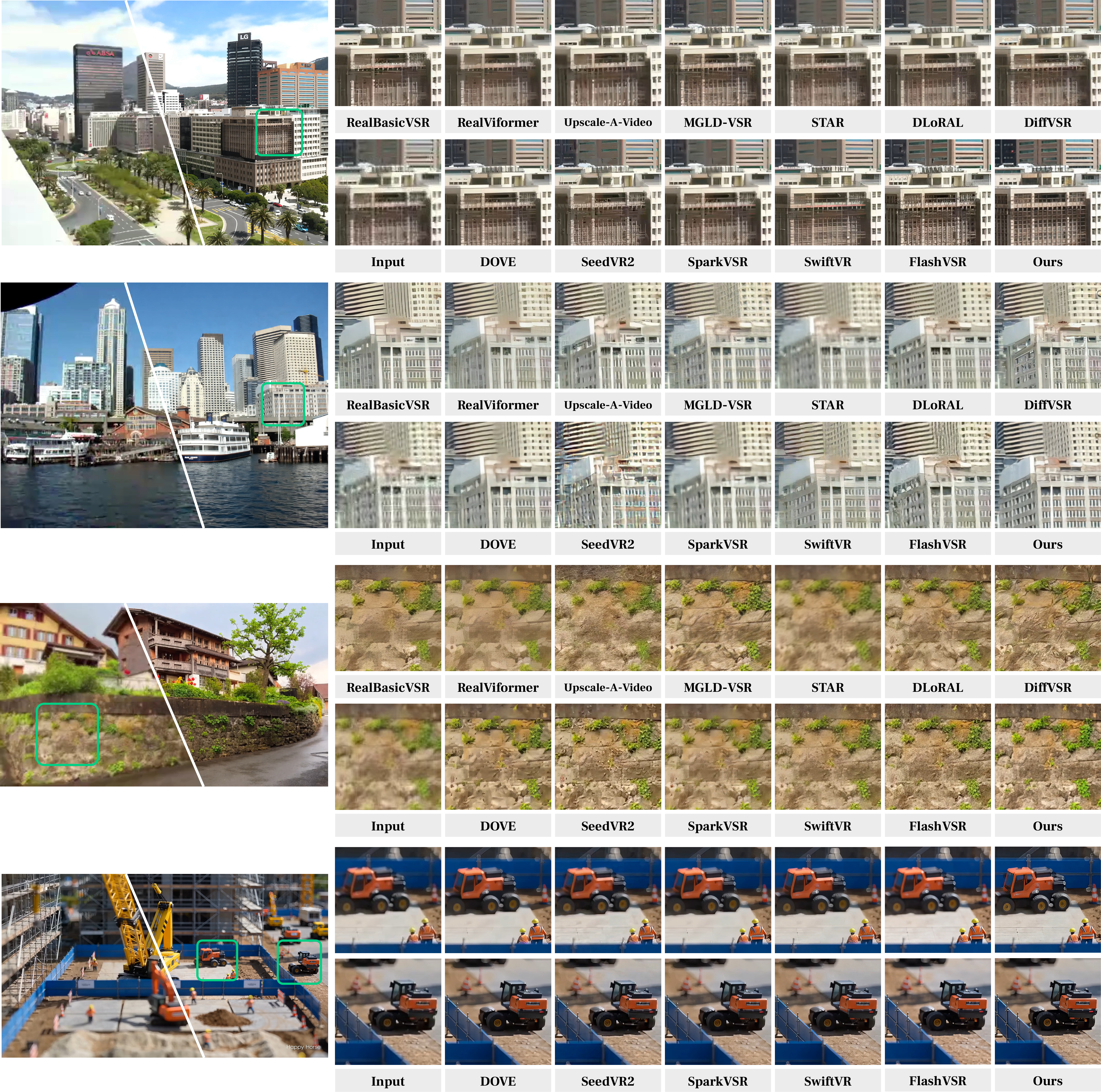}
\caption{Additional visual comparisons. Each example shows the input frame
with a marked region and the corresponding crops from VSR methods.
The scenes include building facades, masonry, and construction machinery.}
\label{fig:app_visual_comparison}
\end{figure}

\paragraph{Additional visual comparisons.}
Figure~\ref{fig:app_visual_comparison}, placed at the end of this appendix,
shows four further VSR examples. The first two focus on repeated window
patterns and building edges, while the remaining scenes feature irregular
masonry and construction machinery. Aligned crops of the same marked regions
make differences in recovered detail and structural consistency visible
across methods. ReCaVSR retains recognizable scene geometry while
reconstructing fine local textures in these examples.

\paragraph{Human evaluation.}
The study included 35 VideoLQ videos and 15 raters. Participants
rated the outputs of DOVE, SeedVR2-3B, SwiftVR, FlashVSR Tiny,
and ReCaVSR using five-point scales for
overall visual quality (MOS-Q), fine-detail quality (MOS-D), and
temporal stability (MOS-T). Table~\ref{tab:app_human_evaluation}
reports the mean scores. ReCaVSR ranks first on all three dimensions,
with MOS-Q, MOS-D, and MOS-T of 3.84, 3.81, and 3.87.
Compared with the strongest baseline, FlashVSR Tiny, these scores are
higher by 0.06, 0.09, and 0.18 points, respectively. The largest
separation is in temporal stability relative to SwiftVR
(3.87 versus 2.98), alongside a 0.66-point difference in
fine-detail quality (3.81 versus 3.15).
\begin{table}[H]
\centering
\caption{Human evaluation on 35 VideoLQ videos with 15 raters.
MOS-Q, MOS-D, and MOS-T
measure overall visual quality, fine-detail quality, and temporal stability,
respectively. Best results are shown in bold.}
\label{tab:app_human_evaluation}
\begingroup
\small
\begin{tabular*}{0.95\textwidth}{@{\extracolsep{\fill}}lccccc@{}}
\toprule
Metric & DOVE & SeedVR2-3B & SwiftVR & FlashVSR Tiny & ReCaVSR \\
\midrule
MOS-Q $\uparrow$ & 3.15 & 3.40 & 3.32 & 3.78 & \textbf{3.84} \\
MOS-D $\uparrow$ & 3.17 & 3.35 & 3.15 & 3.72 & \textbf{3.81} \\
MOS-T $\uparrow$ & 3.62 & 3.29 & 2.98 & 3.69 & \textbf{3.87} \\
\bottomrule
\end{tabular*}
\endgroup
\end{table}

\subsection{Efficiency Across Resolutions}
\label{app:resolution_efficiency}

Table~\ref{tab:app_multires_efficiency} compares full-pipeline throughput
and peak allocated GPU memory from 720p to 4K on 201 input frames using
one NVIDIA A100-80GB. SwiftVR reaches higher throughput:
23.48 versus 21.20 FPS at 1080p and 5.77 versus 5.40 FPS at 4K.
ReCaVSR uses less than half its peak memory at both resolutions
(15.16 versus 31.36\,GB and 27.53 versus 65.26\,GB).
Across the three paired benchmarks, ReCaVSR has lower LPIPS than SwiftVR.
On VideoLQ, its DOVER is 0.5567 versus 0.4957 for SwiftVR
(Table~\ref{tab:main_results}).
Our LR-conditioned FlashDecoder has higher reconstruction PSNR than
SwiftVR's ReAE~\citep{yan2026swiftvr} (36.25 versus 30.98\,dB;
Table~\ref{tab:ablation_decoder}), although ReAE is faster in
decode-only throughput (141.35 versus 91.71 FPS).
Human ratings in Table~\ref{tab:app_human_evaluation} also favor ReCaVSR
over SwiftVR in fine-detail quality (3.81 versus 3.15 MOS-D) and temporal
stability (3.87 versus 2.98 MOS-T).

\begin{table}[H]
\centering
\caption{\textbf{Resolution scaling on one A100-80GB.}
GPU throughput and peak allocated memory for 201 input frames.
Memory is reported in decimal GB.}
\label{tab:app_multires_efficiency}
\begingroup
\small
\setlength{\tabcolsep}{3.5pt}
\renewcommand{\arraystretch}{1.10}
\begin{tabular*}{\textwidth}{@{\extracolsep{\fill}}l*{8}{r}@{}}
\toprule
& \multicolumn{2}{c}{720p} & \multicolumn{2}{c}{1080p}
& \multicolumn{2}{c}{1440p} & \multicolumn{2}{c}{4K} \\
\cmidrule(lr){2-3}\cmidrule(lr){4-5}
\cmidrule(lr){6-7}\cmidrule(l){8-9}
Method & FPS $\uparrow$ & GB $\downarrow$
& FPS $\uparrow$ & GB $\downarrow$
& FPS $\uparrow$ & GB $\downarrow$
& FPS $\uparrow$ & GB $\downarrow$ \\
\midrule
RealViformer & 31.88 & 5.44 & 14.87 & 12.36
& 8.69 & 21.66 & 3.88 & 39.03 \\
DOVE & 1.53 & 30.52 & 0.56 & 41.08
& 0.24 & 57.12 & 0.15$^{\dagger}$ & 42.11$^{\dagger}$ \\
SeedVR2-3B & 2.65 & 75.05 & 0.79$^{\dagger}$ & 73.89$^{\dagger}$
& 0.44$^{\dagger}$ & 74.88$^{\dagger}$
& 0.15$^{\dagger}$ & 61.46$^{\dagger}$ \\
Stream-DiffVSR & 1.76 & 10.24 & 0.85 & 25.92
& 0.47 & 61.67 & 0.17$^{\dagger}$ & 44.39$^{\dagger}$ \\
FlashVSR Tiny & 16.52 & 12.90 & 7.80 & 24.45
& 4.41 & 40.61 & 1.30 & 67.99 \\
SwiftVR & 51.06 & 21.25 & 23.48 & 31.36
& 13.37 & 40.74 & 5.77 & 65.26 \\
\midrule
\textbf{ReCaVSR} & 43.61 & 12.97 & 21.20 & 15.16
& 11.99 & 18.31 & 5.40 & 27.53 \\
\bottomrule
\end{tabular*}
\par\smallskip
\noindent\parbox{\textwidth}{\footnotesize
$^{\dagger}$ Spatial tiling after out-of-memory failure.}
\endgroup
\end{table}

\subsection{Instantiated Cache Route}
\label{app:instantiated_route}

The released inference configuration assigns one fixed historical-access
action to each of the 30 DiT layers, as shown in
Table~\ref{tab:app_exported_route}. Ten layers retain no historical K/V,
while the longer recent and combined scopes occur mainly in the middle
of the network. The action is static, but the retained latent positions
advance with the stream.
\begin{table}[H]
\centering
\caption{Deployed layer-wise historical cache policy. Layer indices are
one-based; reserved slots count latent positions per layer.}
\label{tab:app_exported_route}
\begingroup
\small
\setlength{\tabcolsep}{4pt}
\begin{tabular*}{\textwidth}{@{\extracolsep{\fill}}llrr@{}}
\toprule
Action & DiT layers & Layers & Slots per layer \\
\midrule
$\emptyset$ & 1, 2, 4, 6, 9, 22, 25, 27, 29, 30 & 10 & 0 \\
$W_1$ & 3, 28 & 2 & 1 \\
$W_2$ & 5, 7, 13, 17, 21, 24, 26 & 7 & 2 \\
$W_4$ & 8, 11, 20 & 3 & 4 \\
$A$ & 15 & 1 & 2 \\
$W_2+A$ & 10, 19, 23 & 3 & 4 \\
$W_4+A$ & 12, 14, 16, 18 & 4 & 6 \\
\bottomrule
\end{tabular*}
\endgroup
\end{table}

The schedule reserves 66 layer--latent slots, or 2.2 per layer on average,
versus 120 slots for a uniform $W_4$ policy, a 45\% reduction in reserved
history slots. At $704\times1280$ output
resolution, one historical latent has 880 spatial tokens; with 24 heads
of width 128 and BF16 keys and values, each reserved slot occupies
10.31\,MiB. The corresponding historical K/V buffer capacities are
680.63\,MiB and 1237.50\,MiB, respectively. These figures count reserved
history buffers rather than total GPU memory. When recent and anchor
positions overlap, the visible history is smaller than the reserved
capacity; for example, $W_4+A$ can expose five distinct positions while
reserving six slots.

\Needspace{10\baselineskip}
\section{Limitations and Future Work}
\label{app:limitations}

ReCaVSR exports a fixed per-layer cache route for streaming inference.
This makes the cache footprint predictable, but history allocation cannot
adapt to changes in motion or scene content. Our long-video evaluation uses
single-shot sequences and does not measure behavior across hard scene cuts,
where recycled SR latents and historical K/V may carry stale information.
A natural extension is a causal controller that selects among a small set
of cache routes and resets recurrent states at detected scene boundaries.
Such a design would retain one-step blockwise inference while extending
the method to nonstationary streams.

\end{document}